\documentclass[runningheads]{llncs}

\usepackage{eccv}

\usepackage{eccvabbrv}

\usepackage{graphicx}
\usepackage{booktabs}
\usepackage{colortbl}
\usepackage{multirow}
\usepackage{caption}
\usepackage{subcaption}
\usepackage{wrapfig}
\usepackage[accsupp]{axessibility} 
\usepackage{hyperref}

\usepackage{orcidlink}

\begin{document}

\title{Encapsulating Knowledge in One Prompt} 

\author{Qi Li\inst{1}\orcidlink{0000-0002-3294-2858}\and
Runpeng Yu\inst{1}\orcidlink{0000-0001-6321-9614}\and
Xinchao Wang\inst{1}\thanks{Corresponding author.}\orcidlink{0000-0003-0057-1404}}

\authorrunning{Q. Li et al.}

\institute{National University of Singapore \\
\email{\{liqi,r.yu\}@u.nus.edu,\{xinchao\}@nus.edu.sg}}
\maketitle

\begin{abstract}
In this paper, we propose a new knowledge transfer paradigm called Knowledge in One Prompt (KiOP). This paradigm encapsulates knowledge from various models into a solitary prompt without altering the original models or requiring access to the training data, which enables us to achieve efficient and convenient knowledge transfer in more realistic scenarios. From a practicality standpoint, this paradigm not only for the first time proves the effectiveness of Visual Prompt in data inaccessible contexts, but also solves the problems of low model reusability and high storage resource consumption faced by traditional Data-Free Knowledge Transfer, which means that we can realize the parallel knowledge transfer of multiple models without modifying any source model. Extensive experiments across various datasets and models demonstrate the efficacy of the proposed KiOP knowledge transfer paradigm. Without access to real training data and with rigorous storage capacity constraints, 
it is also capable of yielding considerable outcomes when dealing with cross-model backbone setups and handling parallel knowledge transfer processing requests with multiple (more than 2) models. Code is available at \href{https://github.com/LiQiiiii/Encapsulating-Knowledge-In-One-Prompt}{https://github.com/LiQiiiii/Encapsulating-Knowledge-In-One-Prompt}.

\end{abstract}

\section{Introduction}
\label{sec:intro}
In the past few years, the swift progress of large-scale pre-trained models across diverse fields \cite{brown2020language,dosovitskiy2020image,radford2021learning,bao2021beit,he2022masked,devlin2018bert} has given rise to a multitude of efficient knowledge transfer paradigms \cite{lester2021power,li2021prefix,liu2021p,jia2022visual,bahng2022exploring,yang2022deep,Lian_2022_SSF}. These paradigms are designed to enhance the models' accessibility by making the transfer of knowledge more efficient, thereby extending their benefits to a broader audience. 
Among these paradigms, inspired by the success of prompt learning in NLP \cite{lester2021power,li2021prefix,liu2021p}, the concept of visual prompt \cite{bahng2022exploring} has emerged for tasks in the vision field. Visual prompts involve trainable parameters directly into the input pixel space, facilitating efficient knowledge transfer without altering the pre-trained model. Although it has not yet demonstrated a clear edge over the established knowledge transfer frameworks, many researchers have made considerable effort to unlock its properties and significantly enhance its utility \cite{chen2023understanding,oh2023blackvip,wu2022unleashing,huang2023diversity,li2023exploring,chen2023visual,sohn2023visual,wu2023quantifying,huang2023prompt}. It is expected that Prompt-as-a-Service (PaaS) \cite{wu2023quantifying,huang2023prompt} will gain popularity.

As a new knowledge transfer context, unlike Machine Learning as a Service (MLaaS) \cite{ribeiro2015mlaas,wang2024has}, PaaS allows service providers to offer clients specific prompts. The data of user is combined with the provided prompt and fed into a pre-trained model for predictions, offering a more streamlined and convenient service. Additionally, PaaS providers can leverage a single pre-trained model to handle multiple knowledge transfer service requests which helps to reduce the strain on storage resources and enables more efficient service to a wider range of users. However, owing to privacy concerns or the need for confidentiality, sensitive data typically aren't publicly processed. This renders visual prompts, which necessitate real training dataset access to achieve knowledge transfer, somewhat negligible in such practical scenario.

\begin{figure}[t]
\centering
\includegraphics[width=0.7\columnwidth]{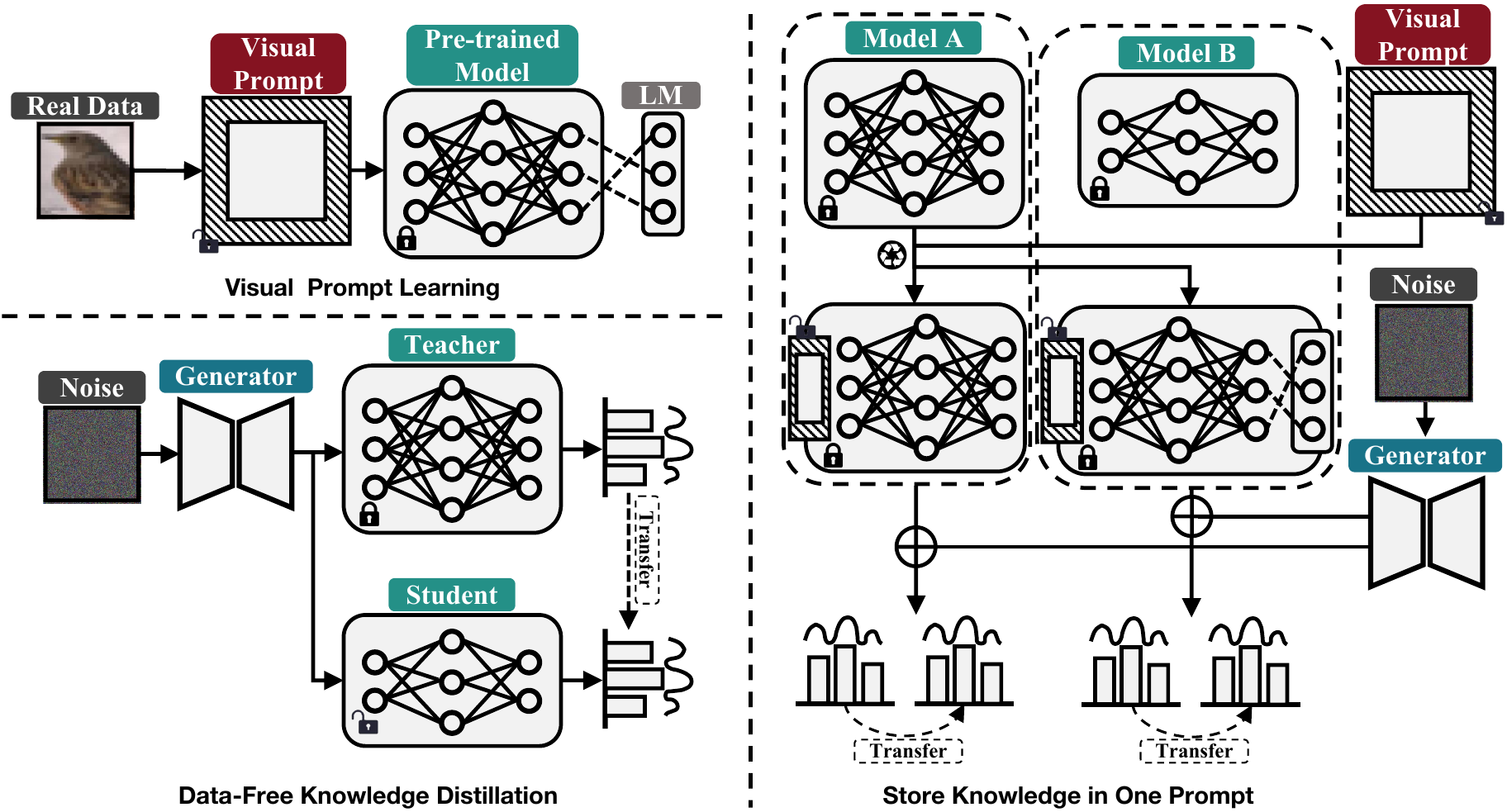} 
\caption{The schematic illustrates Visual Prompt Learning, Conventional Data-Free Knowledge Distillation (DFKD), and our novel Data-free Knowledge Transfer Framework (KiOP). Visual Prompt Learning adjusts pre-trained models for various downstream tasks by appending prompts to inputs while leaving the model frozen. DFKD facilitates knowledge transfer from a teacher model to a student model in the absence of real data. Our KiOP framework ensures effective knowledge transfer without modifications to the pre-trained model or access to real data.}
\label{fig:teaser}
\end{figure}

Furthermore, researchers have thoroughly investigated the effectiveness of knowledge distillation \cite{hinton2015distilling,gou2021knowledge,cho2019efficacy,kim2016sequence,park2019relational,yao2024progressively,yang2022factorizing}—another knowledge transfer framework that is extensively utilized and holds significant potential—in situations lacking training data, which is known as `Data-Free Knowledge Distillation' (DFKD) \cite{li2020learning,fang2019data,micaelli2019zero,lopes2017data,nayak2019zero,chen2019data,yin2020dreaming,ma2020adversarial} and have attained impressive outcomes.
Nevertheless, DFKD comes with three major drawbacks. Firstly, if the student model is already knowledgeable, as mimicking the teacher model's behavior requires updating all its parameters, this process is not only computationally expensive but also risks the knowledge of the student model being catastrophically forgetting \cite{french1999catastrophic,kirkpatrick2017overcoming,goodfellow2013empirical,robins1995catastrophic}. Moreover, DFKD is storage-intensive, as it necessitates maintaining individual copies of the student model for each corresponding teacher. Thirdly, most DFKD methods can not handle multi-model knowledge transfer simultaneously, instead, the requests are handled in turn, similar to a first-come-first-served process, which is especially difficult in data-free scenarios where methods tend to consume much more computing resources. When faced with a large number of knowledge transfer requests, running these methods from scratch for each request is absolutely intolerable.

Merging the benefits of VP's convenience in knowledge storage and lightweight knowledge transfer with the advantages of DFKD in sensitive data protection raises an intriguing possibility: \textit{Whether VP can alleviate the problems of DFKD under the massive storage requirements, the lack of model reusability, and non-parallelism in multi-model knowledge transfer, while drawing intuitions from DFKD to alleviate its dependence on data?}

In this paper, we give an affirmative answer to the above question. We introduce a novel paradigm for knowledge transfer named Knowledge in One Prompt (KiOP). This paradigm balances data autonomy with resource utilization efficiency, confirming the feasibility of encapsulating knowledge from various models within a single prompt. 
Specifically, in our end-to-end training framework, we partition the visual prompt ($\mathcal{VP}$) to more effectively accommodate the varying complexities involved in transferring knowledge from two source models(referred to as model $\mathcal {A}$ and model $\mathcal{B}$ to clarify and avoid ambiguity). Each model is paired with its respective partitions of $\mathcal{VP}$, and each model pair is then processed by a pre-defined synthesis system for data synthesis and storage in their respective data bank. Finally, the two model pairs are jointly trained using the data stored in their corresponding data bank. 
During this process, the parameters of the $\mathcal{VP}$ are updated while the source models are frozen. Consequently, we successfully manage to encapsulate the knowledge from several source models into a single prompt, all without altering the models or requiring access to the actual training data. When new requirements for knowledge transfer are received, since the source model is not modified, we can implement an effective reuse, which alleviates the storage overhead and the catastrophic forgetting problem faced in the traditional DFKD framework, and avoids the security risks caused by real data exposure in the process of visual prompt learning.

\noindent {\bf Contribution.}
We highlight our main contributions as follows. 
We provide a new paradigm called Knowledge in One Prompt (KiOP), which solves the problem of knowledge transfer in a more realistic scenario that was previously unresolved: achieving efficient model reuse and knowledge transfer service parallelism with no access to real data and harsh storage resource constraints. 
We showcase the efficiency and adaptability of KiOP across over 10 dataset pairs, as well as its robustness with various backbones and multi-model (in excess of 2) parallel processing.

\section{Related Work}
\label{sec:RelatedWork}
{\bf Visual Prompt Learning.}
The notion of prompting originated in the realm of Natural Language Processing (NLP) \cite{lester2021power,li2021prefix,liu2021p,zhou2022conditional,zhou2022learning}, with the aim of reconfiguring downstream data to mirror the insights garnered during a model's pre-training phase. 
This process equips frozen pre-trained models with the ability to interpret and adapt to novel tasks more effectively. 
Visual prompt was originally defined in \cite{bahng2022exploring} to imitate the cue idea in NLP, by using input tweaking and label mapping to re-adjust the use of frozen pre-trained vision models for new tasks. Compared with fully fine-tuning, it has the advantage of parameter efficiency and requires much less model storage space.

Recently, many works have explored the effectiveness of $\mathcal{VP}$ in different tasks and application scenarios. In order to further tap the potential of $\mathcal{VP}$, \cite{chen2023understanding} proposed a novel Iterative Label Mapping method to track the dynamics of $\mathcal{VP}$ in the training process, which greatly improves its performance. \cite{oh2023blackvip} explores an adaptive algorithm for $\mathcal{VP}$ when the pre-trained model is a black-box API. \cite{huang2023diversity} proposed to design multiple meta-prompts for a dataset to better adapt to diversity so as to reduce the difficulty of optimization. In the field of security and privacy, \cite{li2023exploring} explores the integration of $\mathcal{VP}$ into canonical differential privacy training methods and proves its effectiveness. \cite{wu2023quantifying} and \cite{huang2023prompt} discussed the vulnerability of $\mathcal{VP}$ to membership inference attacks and backdoor attacks. \cite{li2023towards} for the first time explores the integration capability of $\mathcal{VP}$ on the robustness and generalization ability of the source model when the source model is a robust model. At present, all the works on $\mathcal{VP}$ requires access to the training dataset, the effectiveness of $\mathcal{VP}$ in data-free scenario is still an unexplored field.

\noindent{\bf Data-Free Knowledge Distillation.}
Proposed by \cite{hinton2015distilling}, knowledge distillation is a widely-used knowledge transfer framework that pairs a `teacher' model with a `student' model and manage to distill knowledge from the teacher. This technique has seen applications in various domains \cite{liu2019structured,weinzaepfel2020dope,deng2019relation,chen2021deep,tang2018ranking}. As the field evolves, there is a growing focus on scenarios with greater practical relevance: in real-world situations, the initial training data are not available due to privacy concerns or copyright restrictions. Consequently, the approach of Data-Free Knowledge Distillation has gained popularity. The existing DFKD solutions can be roughly divided into three categories \cite{liu2021data}: Noise Optimization-based, Generative Reconstruction-based and Adversarial Exploration-based. In this work, we mainly draw intuitions from the series of method based on Generative Reconstruction. DAFL \cite{chen2019data} pioneered the use of generative reconstruction in DFKD, introducing two regularization functions concerning activation and prediction to aid in data synthesis.  \cite{ye2020data} suggests a strategy for data-free knowledge amalgamation, aimed at developing a proficient multi-task student network from various single or multi-task teacher networks. 
KegNet \cite{yoo2019knowledge} utilizes a conditional generator along with an auxiliary decoder to distill knowledge from a neural network. \cite{addepalli2020degan} introduces the Data-enriching GAN (DeGAN) framework, which is designed to recover representative samples from a trained classifier utilizing data from a correlated domain. \cite{fang2021contrastive} proposed contrastive model inversion (CMI), which treats data diversity as a goal that can be optimized to mitigate the problem of mode collapse during data reconstruction.

\section{Preliminaries}
\label{sec:pre}
{\bf Knowledge Transfer via Visual Prompt.} $\mathcal{VP}$ is designed to directly add padding, strip, or patch with learnable parameters to the input image. Suppose a $\mathcal{VP}$ parameterized by $\vartheta$ is represented as $\gamma_{\vartheta}$. Then the process of training this $\mathcal{VP}$ can be formalized as follows:
\begin{equation}\label{eq:1}
\vartheta^{*} = \min_{\vartheta} \, \mathbb{E}_{(x_{d},y_{d}) \sim D_{d}} [\mathcal{L}(\mathcal{M}(f_{\psi^{*}}(\gamma_{\vartheta}(x_{d})), y_{d}))] 
\end{equation}
where $\psi^{*}$ represents the weight of the pre-trained model, which is frozen this process; $D_{d}$ represents the distribution of downstream dataset; $\mathcal{M}$ stands for a predefined label mapping (LM) method, which is hard-coded and does not contain learnable parameters. The most commonly used LM methods are Random Label Mapping (RLM) and Iterative Label Mapping (ILM); $\mathcal{L}(\cdot)$ is the loss function for training $\vartheta$, which is generally defined as cross-entropy loss.

\noindent{\bf Data-Free Knowledge Distillation with Data Variety.} 
As previously noted, our focus in this study is mainly on the Generative Reconstruction-based methods within DFKD, since this set of techniques benefits from leveraging the teacher model's prior knowledge as seen in Noise Optimization-based methods, while also informing the enhanced design of the discriminator found in Adversarial Exploration-based methods.

Given that our framework ultimately entails the concurrent learning from two distinct datasets, the variety of synthetic data is especially critical. Hence, we refer to the DFKD module outlined in \cite{fang2021contrastive}. 
This approach takes cues from prior works \cite{wu2018unsupervised,chen2020simple,chen2021exploring} and treat data diversity as an objective that can be optimized
. It employs prior knowledge of batch normalization statistics ($\mathcal{L}_{b}(x)$) from Noise Optimization-based methods, and integrates this with the concept of class priors ($\mathcal{L}_{c}(x)$) and adversarial distillation ($\mathcal{L}_{a}(x)$) to generate synthetic data. This process culminates the first part of the loss function in their method called inversion loss, which is expressed as:
\begin{equation}\label{eq:2}
\mathcal{L}_{inv}(x) = \omega\cdot\mathcal{L}_{b}(x) + \upsilon\cdot\mathcal{L}_{c}(x) + \mu\cdot\mathcal{L}_{a}(x)
\end{equation}

However, the use of the inversion loss alone would lead to the lost of sample diversity. Therefore, they introduce the idea of contrastive learning to model data diversity. The second part of the loss function can be formulated as:
\begin{equation}\label{eq:3}
\mathcal{L}_{cr}(\kappa,\mathcal{D}) = -\mathbb{E}_{x_m \in \kappa} \left[ \log \frac{\exp(sim(x_m, x_m^+, \mathcal{D})/\mu)}{\sum_{n} \exp(sim(x_m, x_n^-, \mathcal{D})/\mu)} \right]
\end{equation}

\noindent where $\kappa$ represents the dataset, and $x_m^+$ represents the positive view of $x_m$ after data augmentation, $x_n^-$ regards all samples except $x_m$ as negative view. $\mu$ stands for the temperature. $sim$ is the cosine similarity which is used to calculate the distinguishable level of image pairs. $\mathcal{D}$ stands for a discriminator which takes the latent features of the image pair as input and distinguish them by projecting them into a new embedding space and calculate the cosine similarity. In addition, a data bank $\mathcal{B}$ is brought in to incrementally store the synthesized data that can be easily distinguished from the historical ones. The final loss function is:
\begin{equation}\label{eq:4}
\min_{\theta_g, z, \mathcal{D}} \left[ \alpha \cdot \mathcal{L}_{cr}(g(z; \theta_g) \cup \mathcal{B}, \mathcal{D}) + \beta \cdot \mathcal{L}_{inv}(g(z; \theta_g)) \right]
\end{equation}
\noindent where $g(\cdot)$ represents the generator parameterized by $\theta_g$, $\alpha$ and $\beta$ represent the weight coefficients of the two terms.

\newcommand{\mysubsection}[1]{\noindent $\bullet$ \textbf{#1}}
\section{Method}

\begin{figure*}[t]
  \centering
  \captionsetup{justification=centering}

  \begin{subfigure}{\textwidth}
    \centering
    \includegraphics[width=\linewidth]{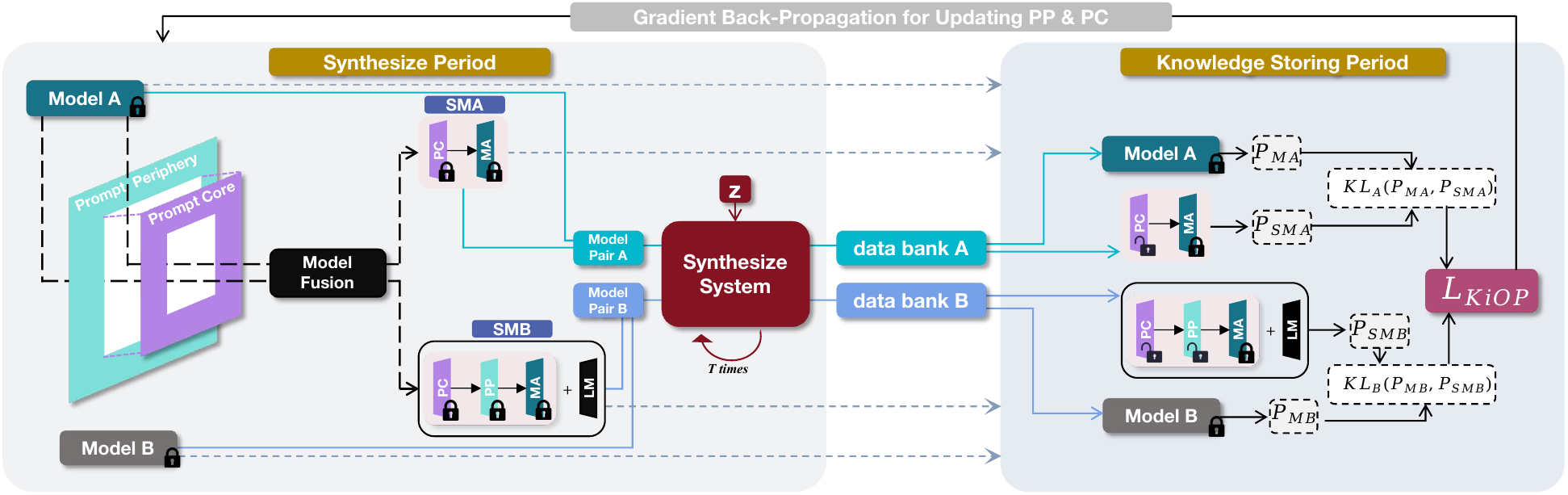}
    \caption{The overall pipeline of the proposed KiOP paradigm.}
    \label{fig:kiop}
  \end{subfigure}

  \begin{subfigure}{.62\textwidth}
    \centering
    \includegraphics[width=\linewidth]{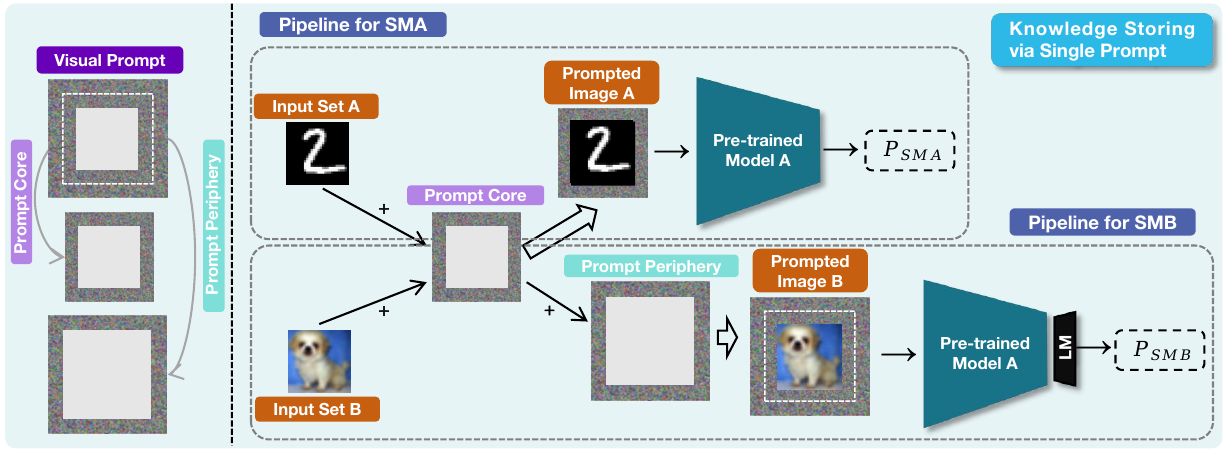}
    \caption{Detailed illustration of the model fusion and input passing logic.}
    \label{fig:pcpp}
  \end{subfigure}%
  \begin{subfigure}{.38\textwidth}
    \centering
    \includegraphics[width=\linewidth]{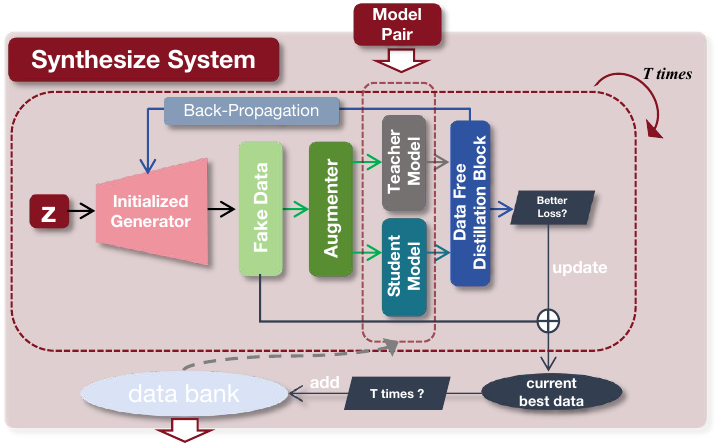}
    \caption{The Synthesize System used to generate synthetic data.}
    \label{fig:ss}
  \end{subfigure}
  \caption{A comprehensive illustration of the proposed paradigm: (a) Overall pipeline, (b) Model fusion and input logic, (c) Synthetic data generation.}
\end{figure*}

Figure \ref{fig:kiop} outlines the KiOP framework, which operates over two main phases: the Synthesize Period and the Knowledge Storing Period. 

\noindent{\bf 4.1. Synthesize Period.} 

\noindent Initially, the system takes in two distinct models and splits the $\mathcal{VP}$ designated for training into two components: the Prompt Core ($\mathcal{PC}$) and the Prompt Periphery ($\mathcal{PP}$). Model $\mathcal{A}$ assumes the role of a secondary transfer agent, serving as a backbone of $\mathcal{VP}$ for the knowledge transfer process. In addition, Model $\mathcal{B}$ is cast as the primary service receiver of the knowledge being transferred to $\mathcal{VP}$. Within our system, both Model $\mathcal{A}$ and Model $\mathcal{B}$ are set frozen, meaning both of them do not going through further training during the whole pipeline. Model $\mathcal{A}$ undergoes a dual Model Fusion operation to yield two derivative models—Student of Model $\mathcal{A}$ ($\mathcal{SMA}$) and Student of Model $\mathcal{B}$ ($\mathcal{SMB}$). $\mathcal{SMA}$ emerges from the fusion of Model $\mathcal{A}$ with $\mathcal{PC}$ and is instrumental in imparting the knowledge of Model $\mathcal{A}$ to $\mathcal{PC}$. Conversely, $\mathcal{SMB}$ is a composite of Model $\mathcal{A}$, $\mathcal{PC}$, and $\mathcal{PP}$. Upon the integration of these elements, a pre-determined label mapping strategy is applied to $\mathcal{SMB}$’s output to align it with the source dataset associated with Model $\mathcal{B}$.

\mysubsection{Model Fusion and Input Passing.} Figure \ref{fig:pcpp} details the components of $\mathcal{SMA}$ and $\mathcal{SMB}$ and the procedure for dividing and utilizing $\mathcal{VP}$. The left side of Figure \ref{fig:pcpp} depicts the division of a full $\mathcal{VP}$ into $\mathcal{PC}$ and $\mathcal{PP}$, intended for later use. This split is strategic, with the goal of appropriately allocating parameter capacity to handle knowledge transfer tasks of varying complexity. For $\mathcal{SMA}$, the challenge is akin to transferring knowledge within the same domain. Although Model $\mathcal{A}$, which we refer to as the core model, has been pre-equipped with a $\mathcal{PC}$ layer, $\mathcal{SMA}$ retains a fundamental grasp of the source data on which the core model was trained, making its task simpler than that of $\mathcal{SMB}$. $\mathcal{SMB}$, sharing the core model with $\mathcal{SMA}$, faces a full-fledged cross-domain learning challenge, as the core model is unfamiliar with Model $\mathcal{B}$'s source data. As shown in Figure \ref{fig:pcpp}, $\mathcal{PC}$ is utilized by both $\mathcal{SMA}$ and $\mathcal{SMB}$, while $\mathcal{PP}$ is used exclusively by $\mathcal{SMB}$. 
In our experiments, we assess the impact of various $\mathcal{PC}$-$\mathcal{PP}$ configurations on overall performance. We discover that different source model pairs respond diversely to the size of $\mathcal{PC}$-$\mathcal{PP}$ setups.

\mysubsection{Synthesize System Design.} Considering a more realistic and challenging data-free scenario, we adopted ideas from previous DFKD works as earlier noted in Sec.\ref{sec:pre} and introduced a Synthesize System. This system is designed to extract synthetic data from two source models to facilitate the learning process. As illustrated in Figure \ref{fig:ss}, the system takes a model pair and a random Gaussian noise vector \( z \) as inputs and uses an initialized generator to create synthetic data. For the design of the Data-Free Distillation Block, we draw inspiration from \cite{fang2021contrastive} as previously stated, which incorporates contrastive learning to enhance the diversity of the synthetic samples, 
thus, we firstly pass the synthesized data through an augmenter for random augmentation, then feed it together with the data previously stored in the data bank to the model pair and get the judgement for its quality, thereby ensuring the diversity of the synthetic data.

\noindent{\bf 4.2. Knowledge Storing Period.} 

\noindent In each training iteration, the two model pairs are passed through the Synthesize System a total of \( T \) times. Note that with each iteration, we start with a freshly initialized generator. As shown in Figure \ref{fig:ss}, during the \( T \) times, once we get a better loss value, we will update the current optimal data.  Upon the completion of \( T \) times, the optimal data is then preserved in the data bank to assist with the training of the prompt in subsequent steps. As the data bank grows with each addition of the best data from these iterations, the prompt is exposed to an increasingly rich array of knowledge. This process ensures a progressively more effective knowledge transfer and storage, as the prompt harnesses a broader spectrum of information with each iteration.

In the Knowledge Storing Period in Figure \ref{fig:kiop}, we train the $\mathcal{PC}$ and $\mathcal{PP}$ with the synthetic data that has been curated in the data bank. It’s important to emphasize that before this period, all the models in the framework—the two source models and one prompt—are set frozen. Focusing on Model Pair $\mathcal{A}$, the aim is to facilitate the knowledge transfer from Model $\mathcal{A}$ to $\mathcal{PC}$ within the $\mathcal{SMA}$ model. To do so, we feed the data from Data Bank $\mathcal{A}$ into both Model $\mathcal{A}$ and $\mathcal{SMA}$. The outputs from both—termed $\mathcal{P_{MA}}$ for the output of Model $\mathcal{A}$, and $\mathcal{P_{SMA}}$ for the output of $\mathcal{SMA}$—are then used to calculate the similarity so that $\mathcal{SMA}$ can imitate the behavior of Model $\mathcal{A}$:
\begin{equation}\label{eq:5}
\mathcal{L}^\mathcal{A}_k = \sum_{\hat{x}_\mathcal{A} \in \mathcal{B}^\mathcal{A}_k} \operatorname{Sim}\left(f_\mathcal{A}(\hat{x}_\mathcal{A}), f_\mathcal{A}(\gamma_{\mathcal{PC,\varphi}}(\hat{x}_\mathcal{A}))\right)
\end{equation}
where $\mathrm{Sim}(\cdot,\cdot)$ is the pre-defined similarity measurement (i.e., KL-Divergance), $\mathcal{B}^\mathcal{A}_k$ represents the state of data bank $\mathcal{A}$ in the $k$-th iteration; $f_\mathcal{A}(\cdot)$ denotes model $\mathcal{A}$; $\hat{x}_\mathcal{A}$ stands for the data selected from $\mathcal{B}^\mathcal{A}_k$; $\gamma_{\mathcal{PC,\varphi}}(\cdot)$ is the Prompt Core parameterized by $\varphi$. For Model Pair $\mathcal{B}$, we use the synthetic data from data bank $\mathcal{B}$ as input for both model $\mathcal{B}$ and $\mathcal{SMB}$, performing identical operations. Due to the pre-defined label mapping method, the output dimension of $\mathcal{SMB}$ matches that of Model $\mathcal{B}$ thus we can measure their similarity as follows:
\begin{equation}\label{eq:6}
\mathcal{L}_k^\mathcal{B} = \sum_{\hat{x}_\mathcal{B} \in \mathcal{B}^\mathcal{B}_k} \operatorname{Sim}\left(f_\mathcal{B}(\hat{x}_\mathcal{B}), \mathcal{M}\left(f_A\left(\gamma_{\mathcal{PP,\eta}}(\gamma_{\mathcal{PC,\varphi}}(\hat{x}_\mathcal{B}))\right)\right)\right)
\end{equation}
where $\mathcal{B}^\mathcal{B}_k$ represents the state of data bank $\mathcal{B}$ in the $k$-th iteration; $f_\mathcal{B}(\cdot)$ denotes model $\mathcal{B}$; $\hat{x}_\mathcal{B}$ stands for the data selected from $\mathcal{B}^\mathcal{B}_k$; $\gamma_{\mathcal{PP}}(\cdot)$ is the Prompt Periphery parameterized by $\eta$; $\mathcal{M}$ is the predefined label mapping method. 

For $\hat{x}_\mathcal{A}$ and $\hat{x}_\mathcal{B}$, they are all generated by
the Synthesize System and stored in the corresponding data bank. For the $k$-th iteration of a Model Pair $j$, this process can be formally described as:
\begin{equation}\label{eq:7}
\mathcal{B}_k^{j} = \mathcal{S}\left(z, (f_{\text{tea}}^{j}(\cdot), f_{\text{stu}}^{j}(\cdot)), k, \mathcal{T}\right)
\end{equation}

where $\mathcal{S}$ represents the Synthesize System as described in Eq.\ref{eq:4}, $(f_{\text{tea}}^{j}(\cdot), f_{\text{stu}}^{j}(\cdot))$ represents the currently used Model Pair, in which $f_{\text{tea}}^{j}(\cdot)$ is the knowledge provider (teacher) of synthetic data, and $\mathcal{T}$ represents predefined round number for $\mathcal{S}$. The final objective function can be defined as:
\begin{equation}\label{eq:7}
\begin{aligned}
\mathcal{L}_{\text{KiOP}} &= \alpha \cdot \mathcal{L}_\mathcal{A}(\mathcal{B}_\mathcal{A}, f_\mathcal{A}, \gamma_{PC,\varphi}, \mathcal{S}) \\
&\quad + \beta \cdot \mathcal{L}_\mathcal{B}(\mathcal{B}_\mathcal{B}, f_\mathcal{A}, \gamma_{PC,\varphi}, \gamma_{PP,\eta}, \mathcal{S}, \mathcal{M})
\end{aligned}
\end{equation}
where $\alpha$ and $\beta$ represent the coefficients that control the weight of the two components. In our experiment, unless stated differently, we set them both to 1.

\section{Experiments}
\noindent{\bf Models and Datasets.} 
To comprehensively evaluate the efficacy of KiOP, we conducted experiments across various Models and datasets. Specifically, we employed backbones from the ResNet series—ResNet-18 and ResNet-50—as well as VGG-13 from the VGG series. For a generalizable assessment, we considered model pairs including (ResNet-18, ResNet-18), (ResNet-50, ResNet-18), (ResNet-18, VGG-13), and (VGG-13, ResNet-18). Moreover, we selected 6 diverse datasets for our experiments: mnist, fashion-mnist (fmnist), cifar10, gtsrb, cifar100, and svhn. 
This breadth ensures our results capture performance across different image types and classification challenges.

\noindent{\bf Settings of Visual Prompt.} 
Without losing generality, we consider using the Padding-type of $\mathcal{VP}$. We default set the size of $\mathcal{VP}$ to $128\times128$.
For the label mapping method, we use the simplest random label mapping (RLM).

\noindent{\bf Baselines and Evaluations.} For a more robust comparison of KiOP's performance, we established multiple baselines:

\mysubsection{Vanilla-KD.} 
It represents the classical DFKD process, in which the student model (Model $\mathcal{A}$ in our case) is not frozen, allowing its parameters to be updated based on the knowledge of the teacher model (Model $\mathcal{B}$) to emulate its behavior. We adopt the DFKD framework as conceptualized in \cite{fang2021contrastive}. Given that the teacher and student datasets may vary in the class numbers, we preserve the weights of the student model's fully connected layer prior to training. When assessing the impact on original performance of the student model, we revert this layer to its initial weights.

\mysubsection{KiOP-Tenuous (KiOP-T).} 
The weights of both models are frozen and a $\mathcal{VP}$ is introduced. However, it is exclusively employed in the knowledge transfer process for Model $\mathcal{B}$: we utilize data from $\mathcal{B}_\mathcal{B}$ as input to $\mathcal{SMB}$, enabling it to mimic the behavior of Model $\mathcal{B}$ and facilitate the transfer of knowledge from Model $\mathcal{B}$ to the $\mathcal{VP}$.

\mysubsection{KiOP-Bilateral (KiOP-B).} 
It requires a single prompt to encapsulate the knowledge from two distinct models—Model $\mathcal{A}$ and Model $\mathcal{B}$. However, the training data for Model $\mathcal{A}$ is available, eliminating the need for $\mathcal{SMA}$ to utilize synthetic data from $\mathcal{B}_\mathcal{A}$ for training purpose. As previously discussed, the $\mathcal{VP}$ here is divided into two segments: the Prompt Core ($\mathcal{PC}$) and the Prompt Periphery ($\mathcal{PP}$). Unless explicitly stated, the default size for $\mathcal{PC}$ is set at $36\times36$, and $\mathcal{PP}$ at $128\times128$.

\mysubsection{KiOP-Bilateral Fake (KiOP-BF).} 
This setting is the most realistic and mainly oriented scenario for KiOP, wherein the training datasets of both models are inaccessible, and a single $\mathcal{VP}$ serves as the repository for the knowledge of both models. $\mathcal{SMA}$ and $\mathcal{SMB}$ are each trained using data from their respective data bank. Despite being a more challenging scenario, empirical evidence suggests that KiOP-BF achieves comparable performance with KiOP-B.

\noindent{\bf 5.1. Performance on Various Difficulty Levels.}

\noindent Table \ref{tab:main} presents the outcomes of experiments conducted across various setups. It can be consistently observed that Vanilla-KD yields the most effective knowledge transfer for Model $\mathcal{B}$, a result that is anticipated given that it leverages the substantial capacity of Model $\mathcal{A}$ itself for learning. Nonetheless, it comes at a significant cost. In the process of learning knowledge from Model $\mathcal{B}$, Model $\mathcal{A}$ experiences a catastrophic forgetting of its original knowledge, akin to the adage of robbing Peter to pay Paul. 
\begin{table*}
\centering
\small
\setlength\tabcolsep{3pt} 
\resizebox{\textwidth}{!}{
\begin{tabular}{c|c|cc|cc|cc|cc|cc|cc}
\toprule
\toprule
\multicolumn{2}{c|}{model settings} & \multicolumn{6}{c|}{Model $\mathcal{A}$: ResNet-18  Model $\mathcal{B}$: ResNet-18} & \multicolumn{6}{c}{Model $\mathcal{A}$: ResNet-50  Model $\mathcal{B}$: ResNet-18} \\
\midrule
\multirow{2}{*}{Dts.$\mathcal{B}$} & metrics & Acc.$\mathcal{B}$ & Acc.$\mathcal{A}$ & Acc.$\mathcal{B}$ & Acc.$\mathcal{A}$ & Acc.$\mathcal{B}$ & Acc.$\mathcal{A}$ & Acc.$\mathcal{B}$ & Acc.$\mathcal{A}$ & Acc.$\mathcal{B}$ & Acc.$\mathcal{A}$ & Acc.$\mathcal{B}$ & Acc.$\mathcal{A}$ \\
\cmidrule(lr){2-14}
&  Dts.$\mathcal{A}$ & \multicolumn{2}{c|}{cifar100} & \multicolumn{2}{c|}{gtsrb} & \multicolumn{2}{c|}{svhn}& \multicolumn{2}{c|}{cifar100} & \multicolumn{2}{c|}{gtsrb} & \multicolumn{2}{c}{svhn} \\
\midrule
\multirow{4}{*}{mnist}
& Vanilla & 57.19\% & 0.15\% & 70.72\% & 1.37\% & 81.70\% & 56.02\% & 61.35\% & 0.15\% & 70.43\% & 6.47\% & 89.29\% & 79.07\% \\
& KiOP-T & 49.40\% & 4.30\% & 48.28\% & 12.60\% & 44.38\% & 16.19\% & 45.42\% & 4.70\% & 56.90\% & 20.90\% & 58.78\% & 16.37\%\\
& KiOP-B & \cellcolor{gray!25}\textbf{44.49\%} & \cellcolor{gray!25}\textbf{37.21\%} & \cellcolor{gray!25}\textbf{46.48\%} & \cellcolor{gray!25}73.05\% & \cellcolor{gray!25}\textbf{42.77\%} & \cellcolor{gray!25}\textbf{86.65\%} & \cellcolor{gray!25}41.14\% & \cellcolor{gray!25}\textbf{43.46\%} & \cellcolor{gray!25}\textbf{51.58\%} & \cellcolor{gray!25}\textbf{78.93\%} & \cellcolor{gray!25}52.40\% & \cellcolor{gray!25}\textbf{87.00\%}\\
& KiOP-BF & \cellcolor{gray!25}43.67\% & \cellcolor{gray!25}36.17\% & \cellcolor{gray!25}45.24\% & \cellcolor{gray!25}\textbf{73.31\%} & \cellcolor{gray!25}39.26\% & \cellcolor{gray!25}83.15\% & \cellcolor{gray!25}\textbf{41.88\%} & \cellcolor{gray!25}41.49\% & \cellcolor{gray!25}48.69\% & \cellcolor{gray!25}78.08\% & \cellcolor{gray!25}\textbf{55.17\%} & \cellcolor{gray!25}83.84\% \\
\midrule
\multirow{4}{*}{fmnist}  
& Vanilla & 44.13\% & 5.38\% & 47.52\% & 5.88\% & 44.41\% & 13.94\%  & 43.31\% & 0.02\% & 48.86\% & 0.77\% & 44.67\% & 16.81\% \\
& KiOP-T  & 43.28\% & 3.56\% & 38.80\% & 14.67\% & 41.28\% & 25.90\%  & 38.88\% & 4.59\% & 42.18\% & 26.03\% & 41.16\% & 22.96\% \\
& KiOP-B  & \cellcolor{gray!25}37.28\% & \cellcolor{gray!25}\textbf{37.40}\% & \cellcolor{gray!25}34.82\% & \cellcolor{gray!25}72.97\% & \cellcolor{gray!25}34.19\% & \cellcolor{gray!25}\textbf{86.58\%}  & \cellcolor{gray!25}\textbf{38.77\%} & \cellcolor{gray!25}\textbf{43.44\%} & \cellcolor{gray!25}39.04\% & \cellcolor{gray!25}\textbf{78.84\%} & \cellcolor{gray!25}38.79\% & \cellcolor{gray!25}\textbf{87.17\%} \\
& KiOP-BF & \cellcolor{gray!25}\textbf{39.02\%} & \cellcolor{gray!25}36.51\% & \cellcolor{gray!25}\textbf{36.47\%} & \cellcolor{gray!25}\textbf{73.64\%} & \cellcolor{gray!25}\textbf{38.02\%} & \cellcolor{gray!25}83.53\% & \cellcolor{gray!25}36.68\% & \cellcolor{gray!25}41.25\% & \cellcolor{gray!25}\textbf{40.39\%} & \cellcolor{gray!25}77.89\% & \cellcolor{gray!25}\textbf{39.86\%} & \cellcolor{gray!25}84.29\% \\
\midrule
\multirow{4}{*}{cifar10}
&Vanilla & 79.55\% & 0.30\% & 79.13\% & 10.92\% & 79.46\% & 5.07\% & 77.60\% & 0.60\% & 77.14\% & 3.07\% & 75.60\% & 4.49\% \\
&KiOP-T  & 44.07\% & 1.13\% & 33.82\% & 6.43\% & 32.80\% & 26.30\% & 44.74\% & 2.18\% & 37.96\% & 14.81\% & 38.77\% & 15.06\% \\
&KiOP-B & \cellcolor{gray!25}\textbf{42.82\%} & \cellcolor{gray!25}\textbf{37.40\%} & \cellcolor{gray!25}28.69\% & \cellcolor{gray!25}73.07\% & \cellcolor{gray!25}27.23\% & \cellcolor{gray!25}\textbf{86.56\%} & \cellcolor{gray!25}\textbf{43.62\%} & \cellcolor{gray!25}\textbf{43.49\%} & \cellcolor{gray!25}\textbf{37.56\%} & \cellcolor{gray!25}\textbf{78.43\%} & \cellcolor{gray!25}37.10\% & \cellcolor{gray!25}\textbf{87.27\%} \\
&KiOP-BF & \cellcolor{gray!25}40.69\% & \cellcolor{gray!25}36.76\% & \cellcolor{gray!25}\textbf{31.24\%} & \cellcolor{gray!25}\textbf{73.86\%} & \cellcolor{gray!25}\textbf{30.26\%} & \cellcolor{gray!25}83.16\% & \cellcolor{gray!25}40.55\% & \cellcolor{gray!25}39.82\% & \cellcolor{gray!25}36.54\% & \cellcolor{gray!25}77.25\% & \cellcolor{gray!25}\textbf{37.21\%} & \cellcolor{gray!25}84.08\% \\
\bottomrule
\bottomrule
\end{tabular}}
\caption{This table presents the outcomes of the KiOP paradigm across various scenarios (under RLM), alongside a comparison with the traditional DFKD. Dts.$\mathcal{A}$ and Dts.$\mathcal{B}$ denote distinct datasets employed in the pre-training of Model A and Model B, respectively. Acc.A and Acc.B are indicators of the performance of $\mathcal{VP}$ in transferring knowledge from the two models. The experiment utilizes two different backbones (ResNet18, ResNet50). Superior results achieved under the KiOP-B and KiOP-BF comparison is highlighted in bold.}
\label{tab:main}
\end{table*}

Subsequently, we delve into various scenarios of KiOP. Initially, we consider a less complex case where the prompt encapsulates the knowledge of a single model (KiOP-T). The outcomes consistently demonstrate the prompt's capability for knowledge transfer. Note that Model $\mathcal{B}$ remains unaltered during this process, thereby preserving its initial knowledge and circumventing the catastrophic forgetting that could ensue from acquiring Model $\mathcal{A}$'s knowledge. Furthermore, in assessing the predictive capabilities of $\mathcal{SMB}$ on dataset $\mathcal{B}$, we observe that although the results are not entirely encouraging, most scenarios exhibit improved performance over Vanilla-KD. For instance, when the dataset pair is (Dts.$\mathcal{B}$, Dts.$\mathcal{A}$) = (mnist, gtsrb), there is a notable increase in accuracy for Acc.$\mathcal{A}$, typically by around 10\%. An outlier to this trend occurs with the pair (Dts.$\mathcal{B}$, Dts.$\mathcal{A}$) = (mnist, svhn), where the Vanilla-KD's results on Acc.$\mathcal{A}$ remain elevated. We hypothesize that this is due to the substantial similarities between the two datasets, such as both being 10-class datasets with classes that represent largely analogous information.

The findings from KiOP-T naturally prompted us to consider the possibility of encapsulating the knowledge from several models within a single prompt.
Results in Table \ref{tab:main} illustrates that a single $\mathcal{VP}$ can adeptly navigate more challenging scenarios, such as KiOP-B and KiOP-BF. Our framework equips the $\mathcal{VP}$ to perform well on Dts.$\mathcal{A}$ while preserving its accuracy on Acc.$\mathcal{B}$. For instance, with cifar100 as Dts.$\mathcal{A}$, our framework elevates Acc.$\mathcal{A}$ from single digits to approximately 40\%. Moreover, with gtsrb and svhn as Dts.$\mathcal{A}$, Acc.$\mathcal{A}$ witnesses a substantial rise from roughly 10–20\% to 70–80\%. We attribute the comparable outcomes between KiOP-BF and KiOP-B to the data-free distillation block utilized in conjunction with our comprehensive end-to-end framework. The former facilitates the generation of high-quality synthetic data from the source model, which in turn enhances the training of the $\mathcal{VP}$, while the latter enables the two sets of synthetic data derived from a unified synthesis system to work better on $\mathcal{VP}$, particularly when both datasets utilize a same $\mathcal{PC}$.

\begin{figure}[htbp]
\centering
\includegraphics[width=0.5\columnwidth]{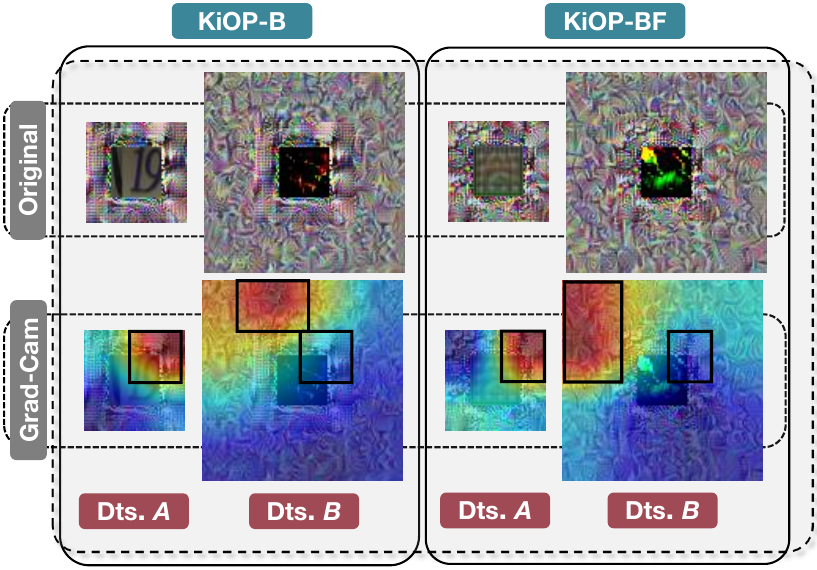}
\caption{$\mathcal{VP}$ and its corresponding Grad-Cam when (Dts.$\mathcal{A}$, Dts.$\mathcal{B}$) = (cifar10, svhn), ResNet-18 as backbone. Size of $\mathcal{PC}$ is set to 64 and $\mathcal{PP}$ is 128.}
\label{fig:grad_cam}
\end{figure}

In order to verify the above conjecture, we visualize the Grad-Cam saliency maps of $\mathcal{VP}$ learned in KiOP-B and KiOP-BF respectively. Results are shown in Figure \ref{fig:grad_cam}. We found that the saliency maps for both KiOP-BF and KiOP-B are remarkably similar, displaying a same pattern: the key areas for $\mathcal{PC}$ and $\mathcal{PP}$ seldom coincide and are predominantly distinct. This suggests that the model responds differently to the two datasets under these conditions, without any adverse interference between them. Furthermore, in the KiOP-BF scenario, the model focuses more intently on the key areas for $\mathcal{PC}$ and $\mathcal{PP}$ with less overlap between them, which implies that the two sets of synthetic data interact more effectively within our end-to-end framework.

\begin{figure}[htbp]
\centering
\includegraphics[width=0.8\columnwidth]{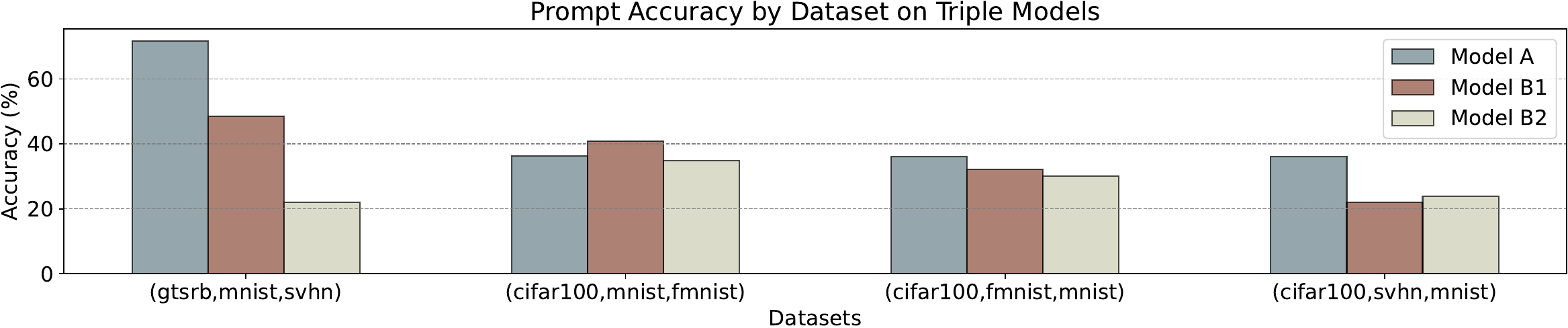}
\caption{VP manages the knowledge storage for three distinct models simultaneously. 
}
\label{fig:triple}
\end{figure}

\mysubsection{KiOP on Multi Models.} Moreover, our investigation extended to the feasibility of KiOP accommodating scenarios involving multiple (more than two) models. In practical scenarios, service providers frequently encounter simultaneous service requests. Therefore, it is worthwhile to investigate the feasibility of processing multiple requests concurrently. Illustrated in Figure \ref{fig:triple} is a case with three original models, designated as Model $\mathcal{A}$, Model $\mathcal{B}_1$, and Model $\mathcal{B}_2$, where Models $\mathcal{B}_1$ and $\mathcal{B}_2$ align with the Model $\mathcal{B}$ configurations depicted in Figure \ref{fig:kiop}. Consequently, the $\mathcal{VP}$ is segmented into three portions sized 36, 128, and 224 respectively, with the first portion being a shared component amongst Model $\mathcal{A}$, Model $\mathcal{B}_1$, and Model $\mathcal{B}_2$, and the second portion shared between Model $\mathcal{B}_1$ and Model $\mathcal{B}_2$. It can be observed that the learned $\mathcal{VP}$ still maintains considerable performance. For instance, with the dataset trio (Model $\mathcal{A}$, Model $\mathcal{B}_1$, Model $\mathcal{B}_2$) = (cifar100, mnist, fmnist), VP achieves an approximate 40\% accuracy for all three models. The comparison between (cifar100, mnist, fmnist) and (cifar100, fmnist, mnist) demonstrates KiOP's robustness to the sequence of $\mathcal{B}_i$ ($i$=1,2) in multi-model scenerios.

\begin{figure}[h!]
\centering
\includegraphics[width=0.75\columnwidth]{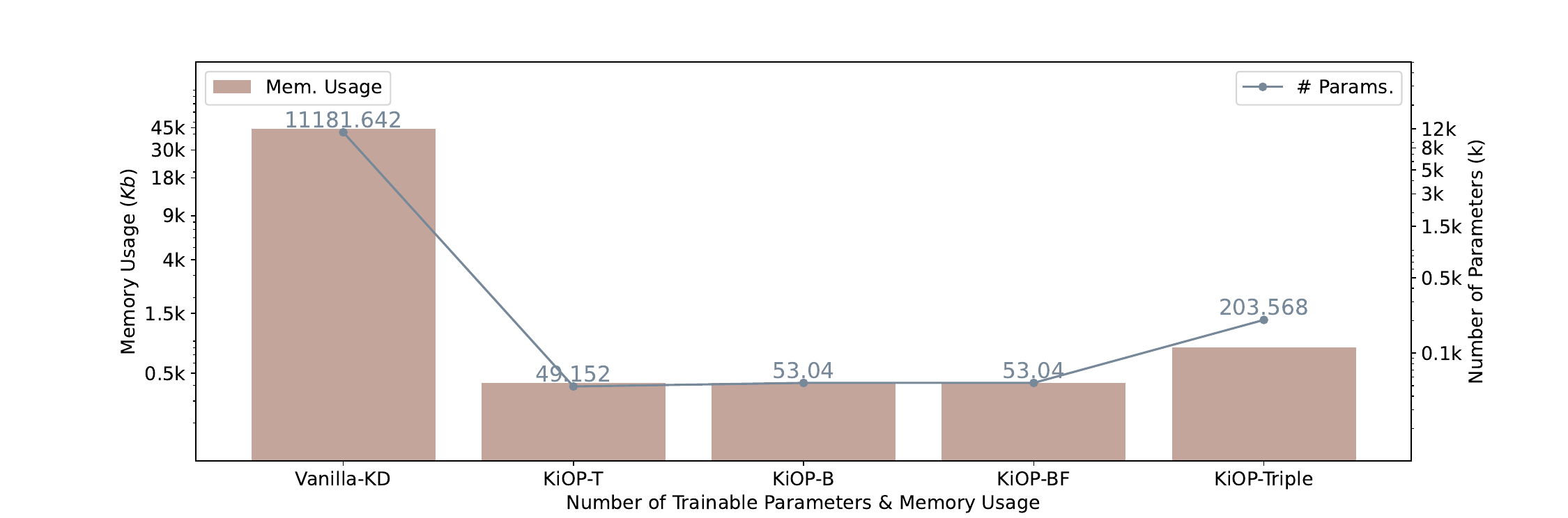}
\caption{KiOP achieves low storage resource usage.}
\label{fig:mems_params}
\end{figure}

\mysubsection{Trainable Parameters and Memory Usage.} 
As shown in Figure \ref{fig:mems_params}, traditional GFKD obliges users to retain the original model, leading to considerable overhead with over 11 million trainable parameters and a memory footprint exceeding 43 MB. In stark contrast, KiOP significantly trims down this requirement by slashing the number of trainable parameters to roughly 50K, thereby substantially reducing the storage resource demands. As a result, KiOP enables users to provide a mere 400 Kb of storage to access knowledge transfer services efficiently.

\noindent{\bf 5.2. Influence of Prompt Core Size.}

\noindent In our earlier experiments, we defaulted to setting the $\mathcal{PC}$ and $\mathcal{PP}$ sizes at 36 and 128, respectively. We now conduct further investigations to ascertain how varying $\mathcal{PC}$ to $\mathcal{PP}$ size ratios influence the ability of $\mathcal{VP}$ to generalize across the two datasets. To this end, we compared four distinct $\mathcal{PC}$ sizes: 36, 48, 64, and 128. The experiment was conducted under the KiOP-B setting. Notably, a $\mathcal{PC}$ size of 128 implies an undifferentiated $\mathcal{VP}$, where the two datasets share the entire $\mathcal{VP}$ without a distinct $\mathcal{PC}$ and $\mathcal{PP}$ division.

\begin{table}
\centering
\small
\setlength\tabcolsep{3pt} 
\resizebox{0.6\columnwidth}{!}{
\begin{tabular}{c|c|cc|cc|cc}
\toprule
\toprule
\multicolumn{2}{c|}{model settings} & \multicolumn{6}{c}{Model $\mathcal{A}$: ResNet-50  Model $\mathcal{B}$: ResNet-18} \\
\midrule
\multirow{2}{*}{Dts.$\mathcal{B}$} & metrics & Acc.$\mathcal{B}$ & Acc.$\mathcal{A}$ & Acc.$\mathcal{B}$ & Acc.$\mathcal{A}$ & Acc.$\mathcal{B}$ & Acc.$\mathcal{A}$\\
\cmidrule(lr){2-8}
&  Dts.$\mathcal{A}$ & \multicolumn{2}{c|}{cifar100} & \multicolumn{2}{c|}{gtsrb} & \multicolumn{2}{c}{svhn}\\
\midrule
\multirow{4}{*}{mnist}			
& 36 & \cellcolor{gray!25}41.14\%& \cellcolor{gray!25}\textbf{43.46\%} & \cellcolor{gray!25}51.58\% & \cellcolor{gray!25}\textbf{78.93\%} & 52.40\% & 87.00\% \\
& 48 & 35.95\% & 26.11\% & \textbf{54.28\%} & 50.23\% & 51.76\%& 73.08\%\\
& 64 & \textbf{41.58\%} & 19.29\%& 53.82\% & 64.16\%& \cellcolor{gray!25}\textbf{52.57\%} & \cellcolor{gray!25}88.92\% \\
&128* & 33.94\% & 11.76\% & 46.66\%& 43.09\% & 41.44\%& \textbf{89.31\%} \\
\midrule
\multirow{4}{*}{fmnist}  		
& 36 & \cellcolor{gray!25}38.77\% & \cellcolor{gray!25}\textbf{43.44\%} & \cellcolor{gray!25}39.04\% & \cellcolor{gray!25}\textbf{78.84\%} & 38.79\% & 87.17\%\\
& 48 & \textbf{39.74\%} & 26.13\%& 37.91\%& 51.29\%& \textbf{42.79\%} & 73.19\% \\
& 64  & 36.21\% & 19.33\% & 36.11\% & 67.98\% & \cellcolor{gray!25}40.92\% & \cellcolor{gray!25}88.76\%\\
& 128* & 36.19\%& 11.56\%& \textbf{40.79\%} & 53.04\%& 24.78\%& \textbf{89.97\%} \\
\midrule
\multirow{4}{*}{cifar10}  	
& 36 & \cellcolor{gray!25}\textbf{43.62}\% & \cellcolor{gray!25}\textbf{43.49\%} & \cellcolor{gray!25}\textbf{37.56\%} & \cellcolor{gray!25}\textbf{78.43\%} & \textbf{37.10\%} & 87.27\%\\
& 48 & 43.59\%& 27.09\%& 36.16\%& 51.73\%& 35.28\%& 73.34\% \\
& 64  & 43.23\% & 21.00\% & 36.01\% & 68.65\% & \cellcolor{gray!25}34.32\% & \cellcolor{gray!25}89.19\%\\
& 128* & 31.24\%& 11.16\%& 34.75\%& 44.44\%& 27.45\%& \textbf{91.31\%} \\
\bottomrule
\bottomrule
\end{tabular}
}
\caption{The effectiveness of \(\mathcal{VP}\) in encapsulating the knowledge from two models varied with the size of the \(\mathcal{PC}\). 
We employed Model \(\mathcal{A}\) with ResNet-50 and Model \(\mathcal{V}\) with ResNet-18 as backbones, 9 different dataset pairs are considered.}
\label{tab:pc_size}
\end{table}
\begin{figure*}[ht!]
    \centering
    \begin{minipage}{\textwidth}
        \centering
        \begin{subfigure}[b]{0.22\textwidth}
            \centering
            \includegraphics[width=\textwidth]{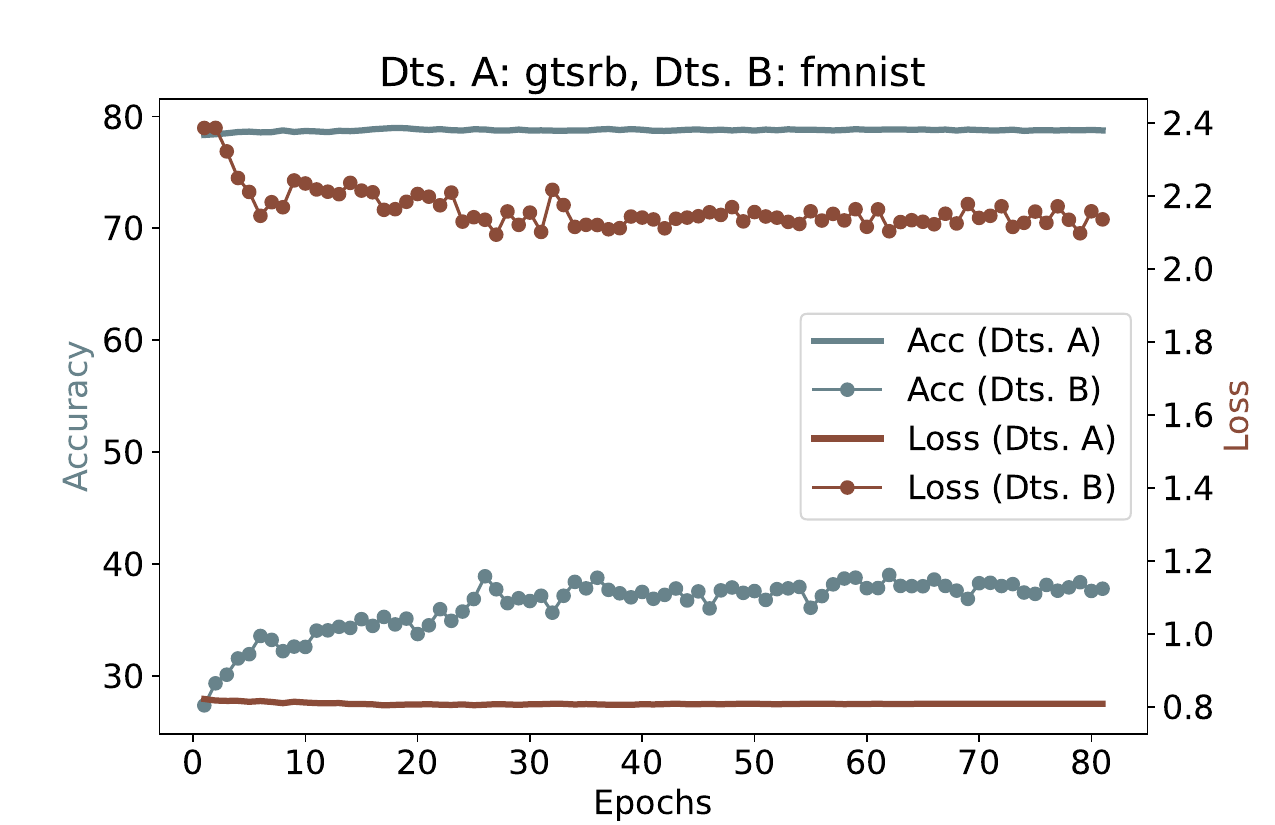}
            \caption{Core size = 36}
            \label{fig:sub1a}
        \end{subfigure}
        \hfill
        \begin{subfigure}[b]{0.22\textwidth}
            \centering
            \includegraphics[width=\textwidth]{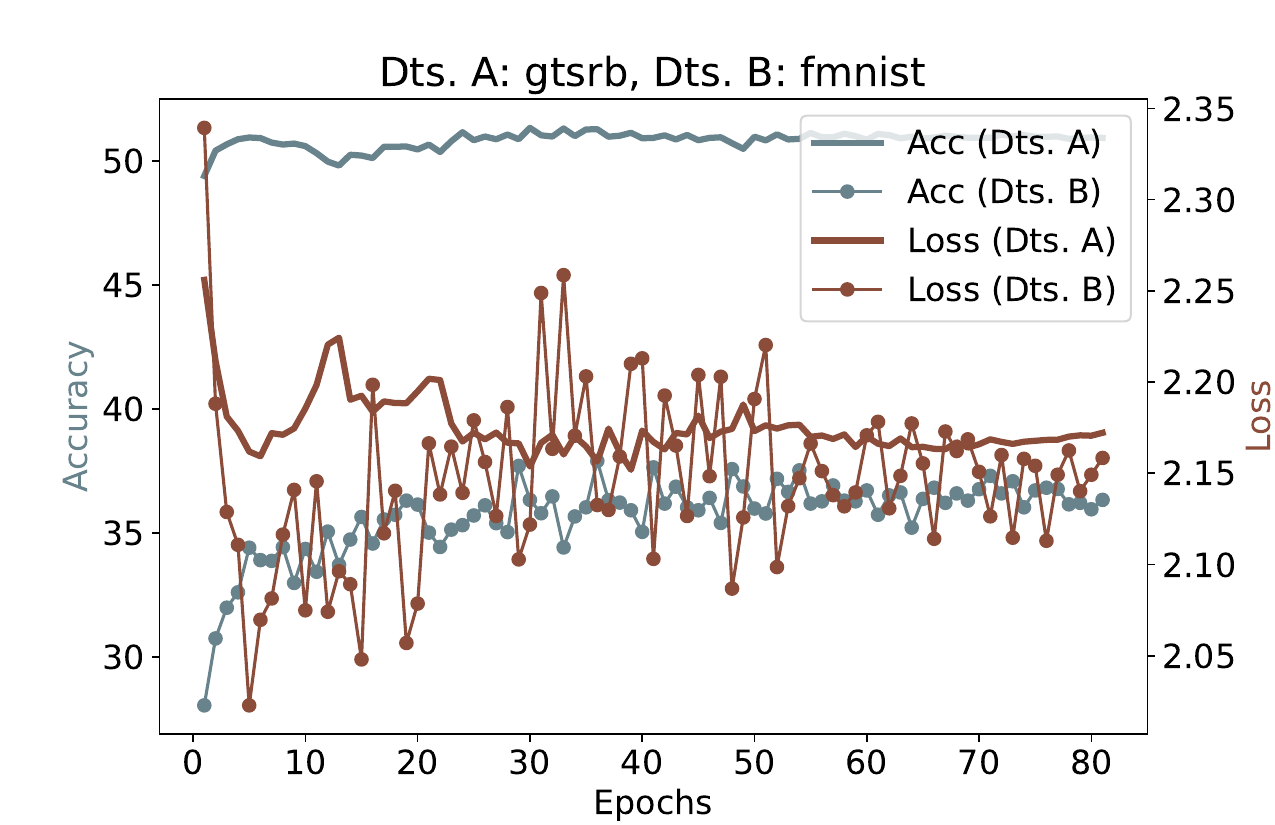}
            \caption{Core size = 48}
            \label{fig:sub2a}
        \end{subfigure}
        \hfill
        \begin{subfigure}[b]{0.22\textwidth}
            \centering
            \includegraphics[width=\textwidth]{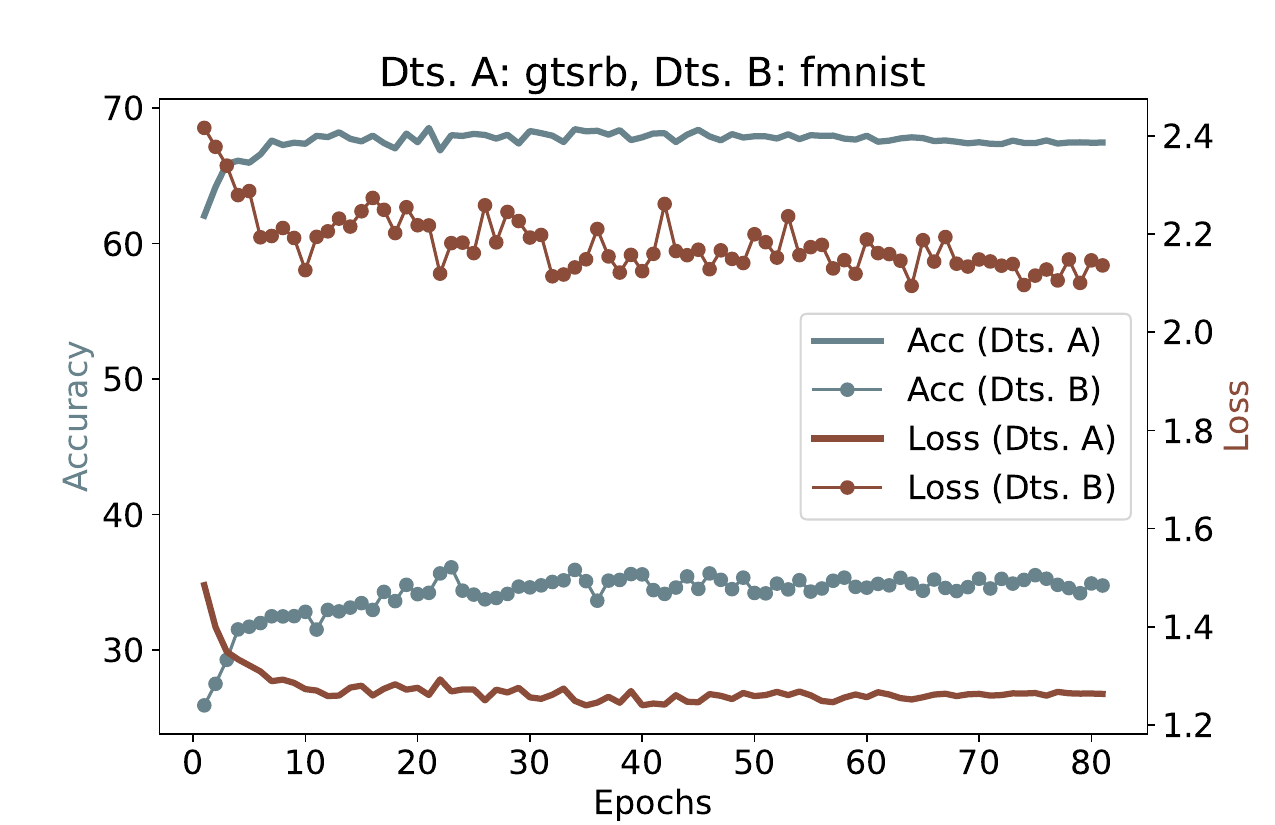}
            \caption{Core size = 64}
            \label{fig:sub3a}
        \end{subfigure}
        \hfill
        \begin{subfigure}[b]{0.22\textwidth}
            \centering
            \includegraphics[width=\textwidth]{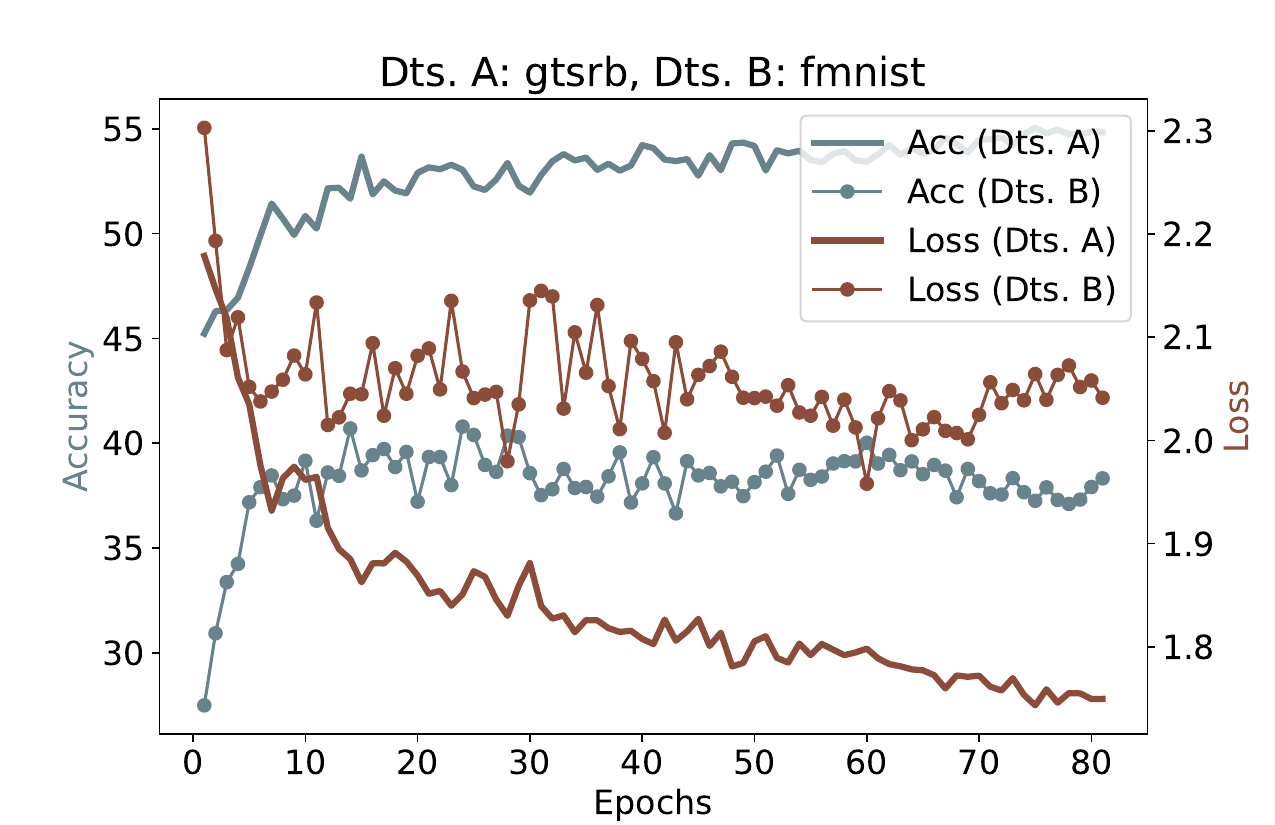}
            \caption{Core size = 128}
            \label{fig:sub4a}
        \end{subfigure}
    \end{minipage}
    \setcounter{subfigure}{0}
        \begin{minipage}{\textwidth}
        \centering
        \begin{subfigure}[b]{0.22\textwidth}
            \centering
            \includegraphics[width=\textwidth]{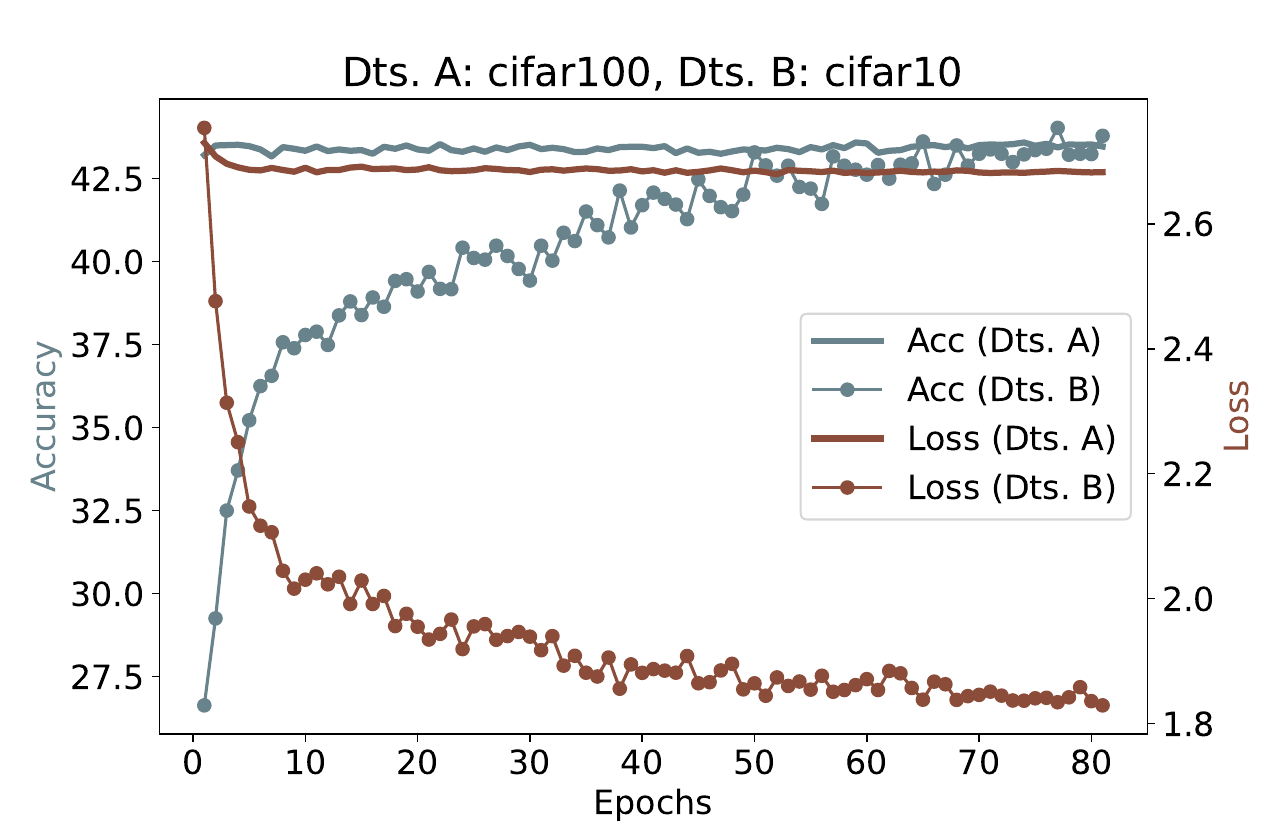}
            \caption{Core size = 36}
            \label{fig:sub1a}
        \end{subfigure}
        \hfill
        \begin{subfigure}[b]{0.22\textwidth}
            \centering
            \includegraphics[width=\textwidth]{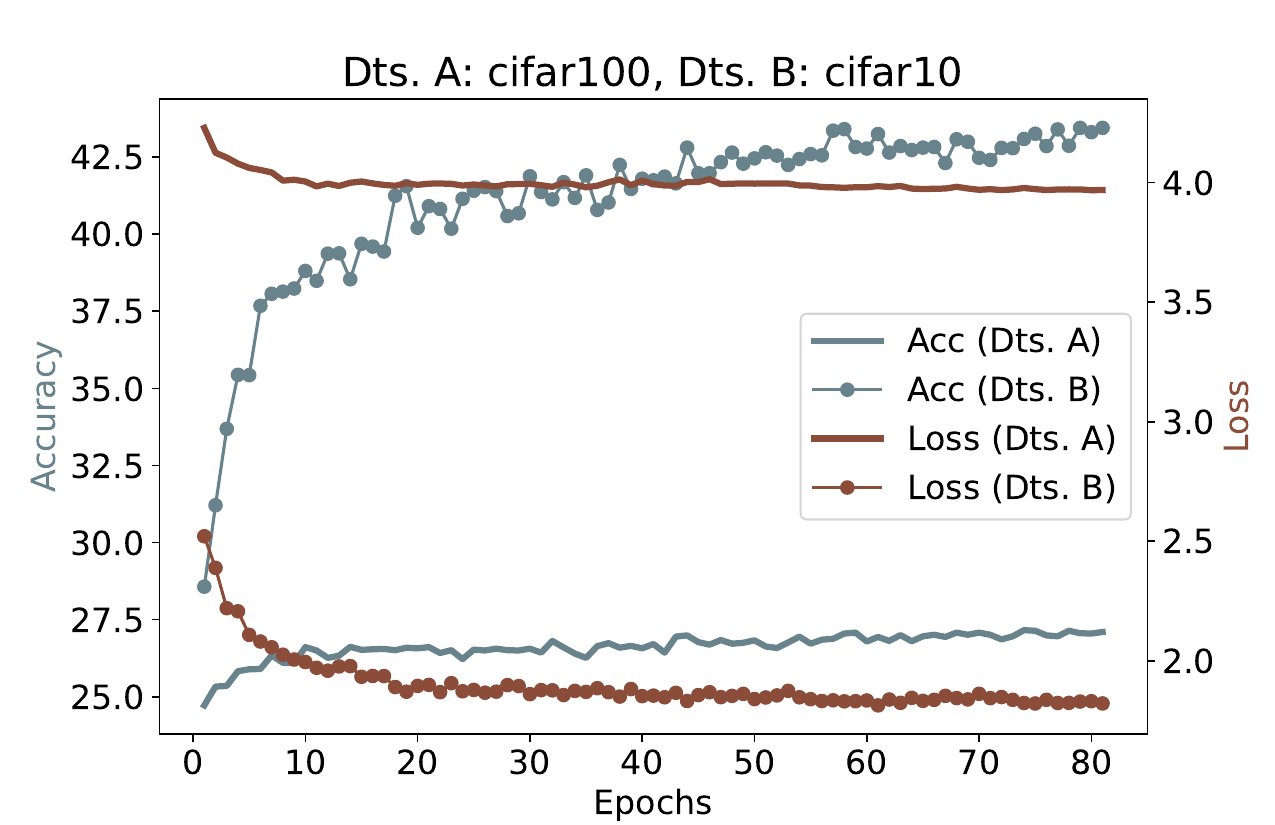}
            \caption{Core size = 48}
            \label{fig:sub2a}
        \end{subfigure}
        \hfill
        \begin{subfigure}[b]{0.22\textwidth}
            \centering
            \includegraphics[width=\textwidth]{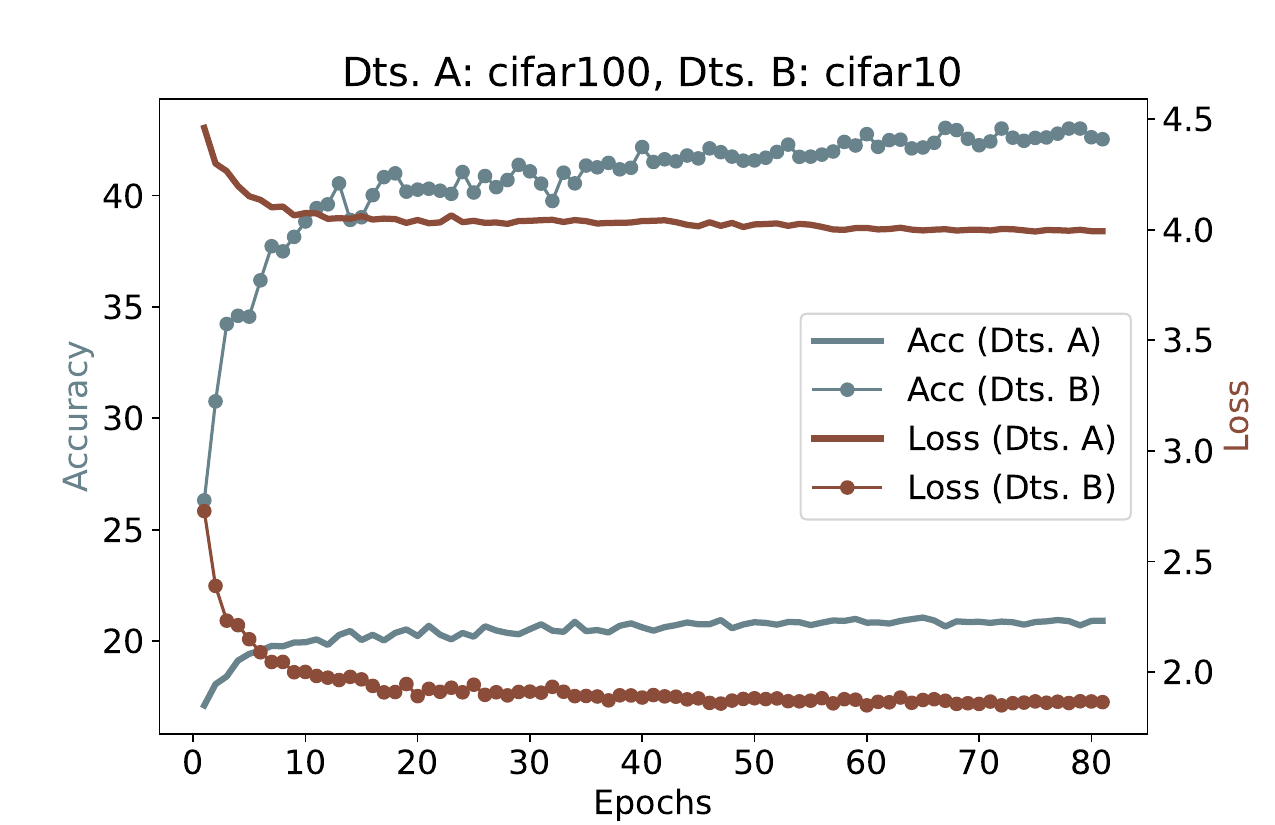}
            \caption{Core size = 64}
            \label{fig:sub3a}
        \end{subfigure}
        \hfill
        \begin{subfigure}[b]{0.22\textwidth}
            \centering
            \includegraphics[width=\textwidth]{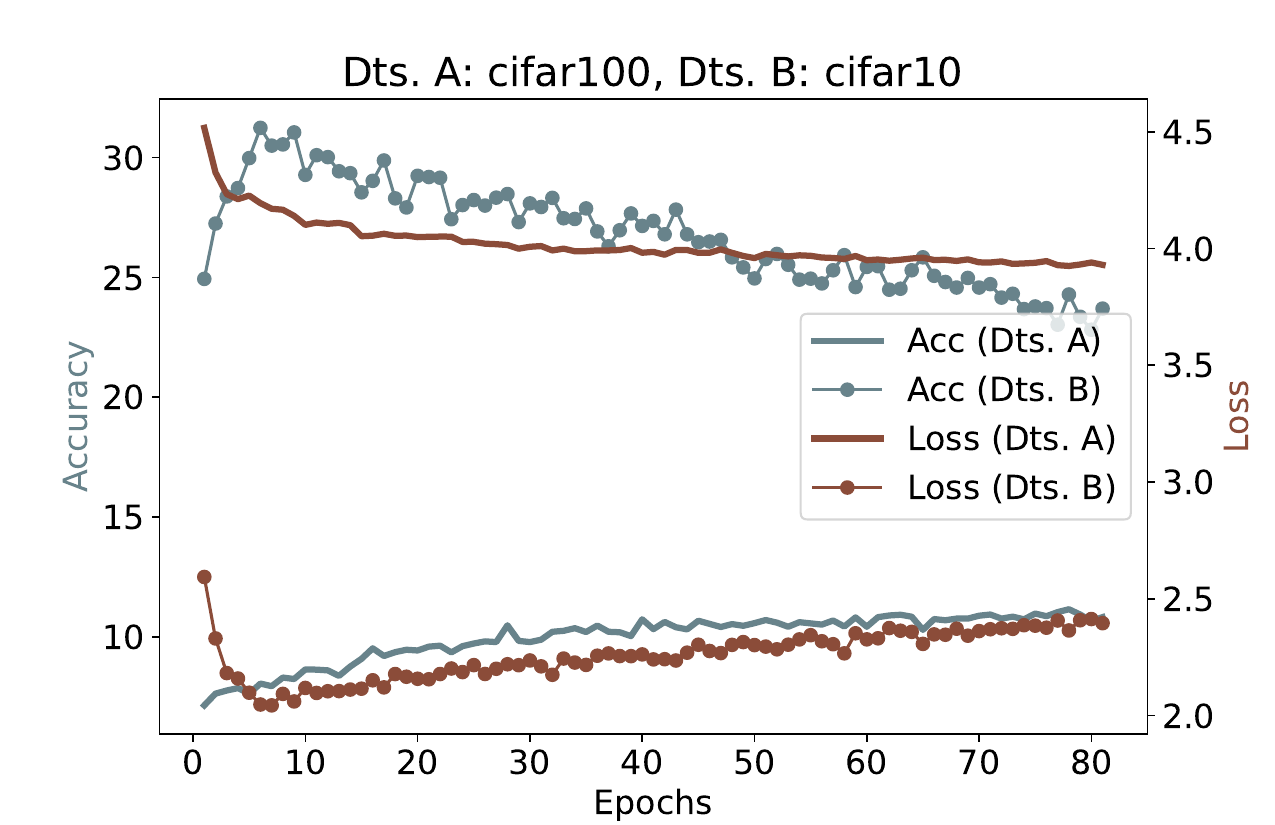}
            \caption{Core size = 128}
            \label{fig:sub4a}
        \end{subfigure}
    \end{minipage}
    \caption{Overall comparison across different datasets and $\mathcal{PC}$ sizes.The first line is the result when (Dts.$\mathcal{A}$, Dts.$\mathcal{B}$) = (gtsrb, fmnist), and the second line is the result when (Dts.$\mathcal{A}$, Dts.$\mathcal{B}$) = (cifar100, cifar10). The size of $\mathcal{PC}$ is set to 36,48,64 and 128.}
    \label{fig:pc_size}
\end{figure*}

Table \ref{tab:pc_size} presents the outcomes for the four aforementioned configurations across 9 distinct dataset pairs. We observed that the performance of the $\mathcal{VP}$ varied with the size of $\mathcal{PC}$ depending on the specific pairing of datasets. For instance, with Dst.$\mathcal{A}$ set to svhn, the $\mathcal{VP}$'s efficacy initially diminishes (36$\rightarrow$48) before it begins to improve (48$\rightarrow$64$\rightarrow$128). Meanwhile, the trends for the corresponding Dts.$\mathcal{B}$ are diverse: for mnist, the first three $\mathcal{PC}$ sizes show negligible differences in performance, but a notable decrease of about 10\% is observed when the $\mathcal{PC}$ size is increased to 128; for fmnist, there's an initial enhancement followed by a dip, with the $\mathcal{PC}$ size of 48 yielding the optimal results; and for cifar10, a continual decline in performance is noted as the $\mathcal{PC}$ size increases. We highlight the best results for each dataset within every dataset pair using boldface, and distinguish the highest overall performance for each dataset pair with a grey background. This clearly demonstrates that, when Dts.$\mathcal{A}$ remains the same, the overall optimal outcomes across different dataset pairs exhibit a consistent trend. Figure \ref{fig:pc_size} depicts the training dynamics with various $\mathcal{PC}$ sizes for the dataset pair (Dts.$\mathcal{A}$, Dts.$\mathcal{B}$) = (gtsrb, fmnist) and (Dts.$\mathcal{A}$, Dts.$\mathcal{B}$) = (cifar100, cifar10). It is observed that a smaller $\mathcal{PC}$ size (36) leads the $\mathcal{VP}$ to prioritize preserving its original performance on Dts.$\mathcal{A}$. Conversely, as the $\mathcal{PC}$ size expands, the VP increasingly focuses on relearning and assimilating knowledge from Dts.$\mathcal{A}$. Regarding Dts.$\mathcal{B}$, as it shares a larger $\mathcal{PC}$ with Dts.$\mathcal{A}$, the training trajectory experiences some fluctuations, yet ultimately stabilizes to reach a satisfactory outcome.

Moreover, an observable trend is that $\mathcal{VP}$ tends to yield superior performance when the data across the two datasets bear resemblance(i.e.,(Dts.$\mathcal{A}$, Dts.$\mathcal{B}$) = (cifar100, cifar10) or (svhn, mnist)). This can be ascribed to the semantic parallels present in the two datasets. Models trained on datasets with similar semantic content tend to develop more comparable knowledge, making it easier for a $\mathcal{VP}$ to assimilate and learn from them.

\noindent{\bf 5.3. KiOP Cross Architectures.}

\begin{table}
\centering
\small
\setlength\tabcolsep{3pt} 
\resizebox{0.7\columnwidth}{!}{
\begin{tabular}{c|c|cc|cc|cc}
\toprule
\toprule
\multicolumn{2}{c|}{model settings} & \multicolumn{6}{c}{Model $\mathcal{A}$: ResNet-18  Model $\mathcal{B}$: VGG-13} \\
\midrule
\multirow{2}{*}{Dts.$\mathcal{B}$} & metrics & Acc.$\mathcal{B}$ & Acc.$\mathcal{A}$ & Acc.$\mathcal{B}$ & Acc.$\mathcal{A}$ & Acc.$\mathcal{B}$ & Acc.$\mathcal{A}$\\
\cmidrule(lr){2-8}
&  Dts.$\mathcal{A}$ & \multicolumn{2}{c|}{mnist} & \multicolumn{2}{c|}{cifar100} & \multicolumn{2}{c}{gtsrb}\\
\midrule
\multirow{2}{*}{cifar10}
& KiOP-B & \textbf{30.15\%} & \textbf{98.30\%} & 36.57\% & \textbf{37.28\%} & 27.84\% & 73.08\% \\
& KiOP-BF & 30.05\% & 97.32\% & \textbf{37.17\%} & 36.36\% & \textbf{28.37\%} & \textbf{73.73\%} \\
\midrule
\multicolumn{2}{c|}{model settings} & \multicolumn{6}{c}{Model $\mathcal{A}$: VGG-13  Model $\mathcal{B}$: ResNet-18} \\
\midrule
\multirow{2}{*}{Dts.$\mathcal{A}$} & metrics & Acc.$\mathcal{B}$ & Acc.$\mathcal{A}$ & Acc.$\mathcal{B}$ & Acc.$\mathcal{A}$ & Acc.$\mathcal{B}$ & Acc.$\mathcal{A}$\\
\cmidrule(lr){2-8}
&  Dts.$\mathcal{B}$ & \multicolumn{2}{c|}{mnist} & \multicolumn{2}{c|}{fmnist} & \multicolumn{2}{c}{cifar10}\\
\midrule
\multirow{2}{*}{gtsrb}  
& KiOP-B  & 39.54\% & 94.18\% & 40.60\% & 93.90\% & 35.30\% & \textbf{94.09\%} \\
& KiOP-BF & \textbf{40.00\%} & \textbf{94.34\%} & \textbf{42.93\%} & \textbf{94.35\%} & \textbf{35.94\%} & 93.56\% \\
\bottomrule
\bottomrule
\end{tabular}
}
\caption{Even with dissimilar backbones, KiOP consistently delivers outstanding results. This experiment was conducted using VGG-13 and ResNet-18. We evaluated 6 diverse dataset pairs. The sizes of $\mathcal{PC}$ and $\mathcal{PP}$ are 36 and 128, respectively.}
\label{tab:cross_arch}
\end{table}
\noindent To further validate the efficacy of KiOP, we expanded our experiments to include different backbone pairs. We specifically tested combinations of ResNet-18 and VGG-13, in both (model $\mathcal{A}$, model $\mathcal{B}$) = (ResNet-18, VGG-13) and (VGG-13, ResNet-18). KiOP's performance was rigorously evaluated across six varied dataset pairs, with the outcomes detailed in Table \ref{tab:cross_arch}. When the backbone pair is reversed to (ResNet-18, VGG-13) with cifar10 as Dts. $\mathcal{B}$, we observed that KiOP's performance remained consistent with previous results under homogeneous backbone conditions in Table \ref{tab:main}. For instance, with (Dts.$\mathcal{A}$, Dts.$\mathcal{B}$)=(gtsrb, cifar10), both Acc.$\mathcal{A}$ and Acc.$\mathcal{B}$ stayed largely the same as those obtained with same backbones. On the other hand, when the model pair is (VGG-13, ResNet-18) with gtsrb as Dts.$\mathcal{A}$, there was a noticeable improvement in Acc.$\mathcal{A}$ by approximately 20\%, while there were gains of about 5\% for Acc.$\mathcal{B}$. These variations likely reflect the complex interplay between differing model backbones and the datasets they are applied to. 

\section{Conclusion}
We propose a new knowledge transfer paradigm named Knowledge in One Prompt (KiOP), which achieves knowledge transfer of n (n$\geq$2) models to a single prompt. Our paradigm eliminates the need for access to real data. 
Moreover, owing to its reliance on training a prompt with a small set of parameters, and its capability to efficiently reuse the original model, it offers significant benefits in terms of storage resource efficiency and the processing of concurrent knowledge transfer requests. 
Sufficient experiments under different setups and different datasets proved its efficiency and adaptability. 
We believe that KiOP can inject new vitality into (data-free) knowledge transfer and makes a substantial move toward facilitating knowledge transfer in serving more realistic application scenarios.

\section*{Acknowledgement}
This work is supported by the Advanced Research and Technology Innovation Centre (ARTIC), the National University of Singapore under Grant (project number: A-0005947-21-00, project reference: ECT-RP2), and the Singapore Ministry of Education Academic Research Fund Tier 1 (WBS: A-0009440-01-00).

\clearpage  

\bibliographystyle{splncs04}
\bibliography{main}

\end{document}